\documentclass{article}

\usepackage[T1]{fontenc}
\usepackage[utf8]{inputenc}
\usepackage[english]{babel}

\usepackage[letterpaper,top=2cm,bottom=2cm,left=3cm,right=3cm,marginparwidth=1.75cm]{geometry}

\usepackage{amsmath}
\usepackage{graphicx}
\usepackage{subcaption}
\usepackage{tikz}
\usetikzlibrary{arrows.meta, positioning}
\usepackage{caption}
\usepackage[colorlinks=true, allcolors=blue]{hyperref}

\title{A Group-Aware Comparative Study of Classical Machine Learning and Deep Learning Methods for Sensor-Based Human Activity Recognition}

\author{Wahyu Pebrianto\thanks{Independent Researcher. Email: \texttt{wahyu.pebrianto1@gmail.com}}}
\date{}  
\begin{document}
\maketitle

\begin{abstract}

Human Activity Recognition (HAR) using wearable or smartphone inertial sensors is commonly evaluated with both classical machine learning and deep learning methods. However, reported performance can be strongly affected by sensor modality, dataset characteristics, and data-splitting strategy. This study presents a controlled comparative evaluation of four classical machine learning models and four deep learning models on two public HAR datasets, UCI-HAR and KU-HAR. Three sensor modalities were considered: accelerometer (ACC), gyroscope (GYRO), and their combination (ACC+GYRO). Classical machine learning models used handcrafted statistical features, while deep learning models operated directly on raw sensor sequences. The primary evaluation employed five-fold StratifiedGroupKFold cross-validation, with subject- or participant-level grouping to prevent group overlap between training and test folds. Macro-F1 was used as the primary performance metric. The results show that deep learning did not consistently outperform classical machine learning. On UCI-HAR, the evaluated deep learning models achieved slightly higher macro-F1 values, whereas on KU-HAR, LightGBM achieved 0.857 compared with 0.828 for the highest-performing evaluated deep learning model. Regarding sensor modality, the combined ACC+GYRO configuration generally outperformed ACC or GYRO alone on both datasets. The evaluation protocol also had a substantial effect: for CNNBiLSTMAttn with ACC+GYRO, random splitting increased macro-F1 by 0.033 on UCI-HAR and 0.093 on KU-HAR compared with grouped evaluation. These findings highlight the importance of model family, sensor modality, dataset characteristics, and group-aware evaluation when assessing sensor-based HAR systems.

\vspace{0.5em}
\noindent\textbf{Keywords:} Human Activity Recognition; Sensor-based HAR; Classical Machine Learning; Deep Learning; Sensor Modality; Grouped Cross-Validation; Macro-F1.
\end{abstract}

\section{Introduction}

\subsection{Background}
Sensor-based Human Activity Recognition (HAR) commonly uses body-worn or mobile inertial sensors, particularly accelerometers and gyroscopes, to capture movement patterns during daily activities~\cite{Lara2013Survey, Bulling2014tutorial}. These approaches have been widely studied because inertial sensing provides a practical way to recognize activities without relying on camera-based observation~\cite{Lara2013Survey, Bulling2014tutorial}. Sensor-based activity recognition has also been investigated in real caregiving environments. Inoue et al.~\cite{Inoue2019Integrating} integrated activity recognition with nursing care records using smartphone sensor data collected in a nursing care facility, demonstrating the practical use of sensor-based HAR in elderly care. Recent work has further investigated location-aware deep learning models for smartphone-based activity recognition in elderly populations, addressing the effect of varying device carrying positions~\cite{Shahzad2026Location}. Traditional sensor-based HAR commonly follows a pipeline consisting of signal acquisition, preprocessing, feature extraction, and classification~\cite{Bulling2014tutorial}. Classical machine learning methods typically operate on handcrafted statistical, time-domain, or frequency-domain features. Early work by Bao et al.~\cite{Bao2004Activity} demonstrated the use of acceleration-derived features with conventional classifiers for activity recognition. Smartphone accelerometers were also used in early mobile activity-recognition systems for daily activities, demonstrating the practicality of mobile sensing for HAR~\cite{Kwapisz2011Activity}. Public benchmarks such as UCI-HAR subsequently provided standardized smartphone-based inertial data and activity labels for evaluating HAR algorithms~\cite{Anguita2013Public}.

With the development of deep learning~\cite{LeCun2015Deep}, researchers began to learn representations directly from raw sensor sequences instead of relying entirely on handcrafted features~\cite{Hammerla2016Deep}. Convolutional neural networks (CNNs) can learn local patterns from sensor signals~\cite{Raj2023improved}, while recurrent neural networks such as long short-term memory (LSTM) models can capture temporal dependencies~\cite{Hammerla2016Deep}. Hybrid CNN-LSTM architectures combine these capabilities by using convolutional layers for feature extraction and recurrent layers for temporal modeling~\cite{Ordonez2016Deep}. Recurrent deep learning approaches have also been investigated for learning temporal representations directly from wearable sensor data~\cite{Pawlyta2020Deep}. Although both classical and deep learning approaches have demonstrated strong performance, their relative performance can vary across datasets and experimental settings. Previous studies have highlighted the influence of model architecture and hyperparameters~\cite{Hammerla2016Deep}, feature extraction~\cite{Rehman2024Human}, and sensor fusion~\cite{Ordonez2016Deep} on HAR performance. Therefore, direct comparison under a common experimental protocol is important for understanding the practical differences between these approaches.

\subsection{Problem Statement}
Three issues motivate this study. First, it is not always clear whether deep learning provides a consistent advantage over classical machine learning for sensor-based HAR. Deep learning models can learn representations directly from raw signals, but feature-based classical machine learning models may remain competitive, particularly when informative features can be extracted effectively.

Second, the usefulness of individual sensor modalities may vary. Accelerometer and gyroscope signals capture different aspects of motion, and combining them may provide complementary information. However, the benefit of sensor fusion may depend on the activity, sensor configuration, feature representation, and learning method~\cite{Ordonez2016Deep, Shoaib2014Fusion}.

Third, the evaluation protocol can substantially affect reported HAR performance. Random sample-based splitting can place samples from the same participant in both training and test sets, which may lead to optimistic estimates of generalization. Recent comparative evidence in wearable HAR shows that random $k$-fold validation can yield substantially higher performance than Leave-One-Subject-Out (LOSO) validation because samples from the same subjects may appear in both training and test folds~\cite{Rehman2024Human}.

These issues make it difficult to interpret model comparisons when model family, sensor modality, and data-splitting strategy are not controlled.

\subsection{Objective, Contributions, and Scope}

This study aims to systematically compare classical machine learning and deep learning methods for sensor-based HAR under a common group-aware evaluation framework. Specifically, the study investigates: (i) the performance of classical machine learning and deep learning models under grouped evaluation with subject- or participant-level grouping; (ii) the effect of accelerometer (ACC), gyroscope (GYRO), and combined ACC+GYRO modalities; and (iii) the effect of grouped versus random data splitting on reported HAR performance. Two public HAR datasets, UCI-HAR~\cite{Anguita2013Public} and KU-HAR~\cite{Sikder2021KU}, are used to examine whether the observed patterns are consistent across datasets with different numbers of activity classes and dataset characteristics. The main contributions of this study are:
\begin{enumerate}
    \item A controlled comparison of four classical machine learning methods and four deep learning architectures for sensor-based HAR.
    \item A sensor modality comparison across ACC, GYRO, and ACC+GYRO using a consistent evaluation framework.
    \item An empirical analysis of grouped and random splitting to quantify the effect of the evaluation protocol on reported performance.
    \item An analysis of classification errors and computational cost to complement the overall performance comparison.
\end{enumerate}

This study does not propose a new architecture and does not aim for state-of-the-art results. All evaluated model configurations use fixed, non-tuned hyperparameters and are applied consistently across the evaluated dataset--modality combinations. The primary contribution is a controlled comparison under a group-aware evaluation protocol designed to prevent group overlap between training and test folds, rather than a claim that one model family is universally superior.

\section{Related Work}

\subsection{Machine Learning for Sensor-based HAR}
Machine learning has been widely used in sensor-based HAR, particularly through handcrafted feature extraction. Typical approaches calculate statistical and signal-processing features from segmented sensor windows and then use classifiers such as logistic regression, support vector machines, decision trees, or ensemble methods~\cite{Lara2013Survey, Bulling2014tutorial}. Bao et al.~\cite{Bao2004Activity} provided an early example of activity recognition using acceleration features and conventional classification methods. Inoue et al.~\cite{Inoue2019Integrating} further demonstrated the use of statistical sensor features with an Extremely Randomized Tree classifier in a real nursing-care setting.

The UCI-HAR dataset~\cite{Anguita2013Public} further established a widely used benchmark for smartphone-based HAR. It contains inertial measurements from accelerometers and gyroscopes together with activity labels and subject identifiers, enabling both algorithm development and subject-based evaluation~\cite{Anguita2013Public}. Feature-based machine learning remains relevant because carefully designed statistical and frequency-domain representations can provide compact inputs for conventional classifiers. Sensor fusion studies have also examined how accelerometer and gyroscope information can be used individually or in combination for activity recognition~\cite{Shoaib2014Fusion}. More recent comparative research also indicates that feature-based models can remain competitive under subject-independent validation, highlighting the importance of feature representation in HAR evaluation~\cite{Rehman2024Human}.

\subsection{Deep Learning for Sensor-based HAR}
Deep learning~\cite{LeCun2015Deep} methods were introduced to reduce reliance on handcrafted features by learning representations directly from sensor signals. Recent surveys have systematically documented the development of deep learning methods for sensor-based HAR, including convolutional, recurrent, and other deep architectures~\cite{Wang2019Deep}. Hammerla et al.~\cite{Hammerla2016Deep} systematically compared convolutional and recurrent approaches across multiple wearable-sensor datasets and showed that the relative suitability of different architectures depends on the characteristics of the HAR task. CNN-based architectures have been applied to sensor-based HAR using raw accelerometer data~\cite{Raj2023improved, Zeng2014Convolutional}, while deeper CNNs have been developed for multichannel inertial time-series HAR to learn higher-level representations from raw sensor signals~\cite{Yang2015Deep}. Transfer learning from pre-trained CNN architectures such as ResNet50V2 has also been explored for elderly activity classification using accelerometer-derived heatmaps~\cite{Jana2026Sensor}.

Recurrent models have been studied independently and in combination with convolutional frontends for sensor-based activity recognition. Inoue et al.~\cite{Inoue2018Deep} developed a deep recurrent neural network for mobile HAR using raw accelerometer data, while Murad et al.~\cite{Murad2017Deep} investigated deep recurrent architectures for learning temporal representations from sequential sensor data. Domain-specific applications such as skiing recognition have further demonstrated the adaptability of recurrent architectures to activity-recognition tasks beyond standard daily-activity benchmarks~\cite{Pawlyta2020Deep}. CNN-LSTM architectures have also been investigated to combine convolutional feature extraction with recurrent temporal modeling. Ordóñez et al.~\cite{Ordonez2016Deep} applied CNN-LSTM to multimodal wearable activity recognition, while Sharma et al.~\cite{Sharma2022Human} applied a CNN-LSTM network to smartphone sensor data for human activity recognition. Hybrid architectures such as DeepSense have further integrated convolutional and recurrent networks to model heterogeneous mobile sensing data~\cite{Yao2017DeepSense}, while comparative studies have evaluated combinations of classical and deep learning models for activity recognition and postural transition detection~\cite{Chadha2024Hybrid}.

Beyond these core architectures, research has also explored sensor fusion and more advanced representation-learning strategies. Multi-sensor fusion has been investigated to exploit complementary information from different inertial sensors~\cite{Shoaib2014Fusion}. Attention mechanisms have been introduced to assign different weights to informative components of HAR sequences~\cite{Mahmud2020Human}. Recent reviews have further summarized the development, applications, and challenges of deep learning for wearable-sensor HAR~\cite{Zhang2022Deep}. More recent work has addressed user generalization in wearable HAR through multi-task contrastive learning and cross-user evaluation~\cite{Guo2025Towards}. In parallel, transformer-based and non-transformer-based HAR architectures have been systematically examined, including their computational trade-offs~\cite{Leite2024Transformer}. These developments motivate the CNN, BiLSTM, CNN-LSTM, and attention-pooled architectures evaluated in the present study.

\subsection{Research Gap and Study Positioning}
Previous studies have compared classical machine learning and deep learning approaches for sensor-based HAR. For example, Balaha and Hassan~\cite{Balaha2023Comprehensive} conducted a broad comparison of machine learning and deep learning approaches on WISDM and UCI-HAR, including feature-based machine learning and deep architectures such as 1D-CNN, BiLSTM, and GRU. The present study differs by evaluating predefined model configurations within a common group-aware cross-validation framework and by explicitly examining sensor modality and grouped versus random splitting.

The evaluation protocol is particularly important. Recent comparative evidence in wearable HAR shows that random $k$-fold validation can yield substantially higher performance than Leave-One-Subject-Out validation because samples from the same subjects may appear in both training and test folds~\cite{Rehman2024Human}. Data heterogeneity across subjects, sensor modalities, and environmental conditions has also been identified as an important challenge in sensor-based HAR~\cite{Ye2024Machine}.

The present study therefore performs a controlled comparative evaluation in which classical feature-based models and raw-sequence deep learning models are evaluated under the same group-aware cross-validation framework. The study further examines ACC, GYRO, and ACC+GYRO modalities and explicitly compares grouped and random splitting. This positioning is also relevant to KU-HAR, an open HAR dataset containing smartphone accelerometer and gyroscope recordings across 18 activity classes~\cite{Sikder2021KU}. In the present experiments, KU-HAR is processed from the raw time-domain recordings, while UCI-HAR is used as a second benchmark for comparison~\cite{Anguita2013Public, Sikder2021KU}.

\section{Materials and Methods}

\subsection{Datasets}
Two public sensor-based HAR datasets, UCI-HAR~\cite{Anguita2013Public} and KU-HAR~\cite{Sikder2021KU}, were used to evaluate the classical machine learning and deep learning methods considered in this study. The datasets were selected because they differ in the number of activities, sensor configuration, sampling frequency, and data volume, allowing the robustness of the comparison to be examined under different HAR settings. Table~\ref{tab:datasets} summarizes the main characteristics of the two datasets.

\subsubsection{UCI-HAR}
The UCI-HAR dataset contains smartphone inertial measurements from 30 subjects performing six activities: WALKING, WALKING\_UPSTAIRS, WALKING\_DOWNSTAIRS, SITTING, STANDING, and LAYING~\cite{Anguita2013Public}. The dataset provides nine sensor channels consisting of total acceleration, body acceleration, and body gyroscope measurements. The original data are provided as fixed-length windows of 128 samples at 50\,Hz, corresponding to 2.56\,s, with 50\% overlap. In the present study, the original windows were used without additional window generation. The UCI-HAR release also provides a pre-computed 561-feature vector for each window~\cite{Anguita2013Public}. These features were not used in this study. Instead, handcrafted features were recomputed from the sensor channels using the common 17-feature-per-channel procedure described in Section~\ref{sec:models}. This choice ensured that UCI-HAR and KU-HAR used the same handcrafted feature-extraction recipe across the evaluated sensor modalities. The resulting dataset contains 10{,}299 windows and 30 subject groups. The original train/test split was retained only as a secondary evaluation for comparison with published UCI-HAR results, while the main experiments used grouped cross-validation with subjects as the grouping unit.

\subsubsection{KU-HAR}
KU-HAR~\cite{Sikder2021KU} contains 18 human activity classes recorded using smartphone accelerometer and gyroscope sensors. In this study, the experiments used the raw time-domain recordings from the public release (folder \texttt{1.Raw\_time\_domian\_data}), rather than the anonymized \texttt{Time\_domain\_subsamples.csv}. Each raw recording contains three accelerometer and three gyroscope channels. The recordings were segmented into non-overlapping windows of 300 samples, corresponding to 3\,s at 100\,Hz. The participant identifier was obtained from the recording filename and used as the grouping variable, preventing data from the same participant from appearing in both training and test folds.

\begin{figure}[htbp]
\centering
\includegraphics[width=\textwidth]{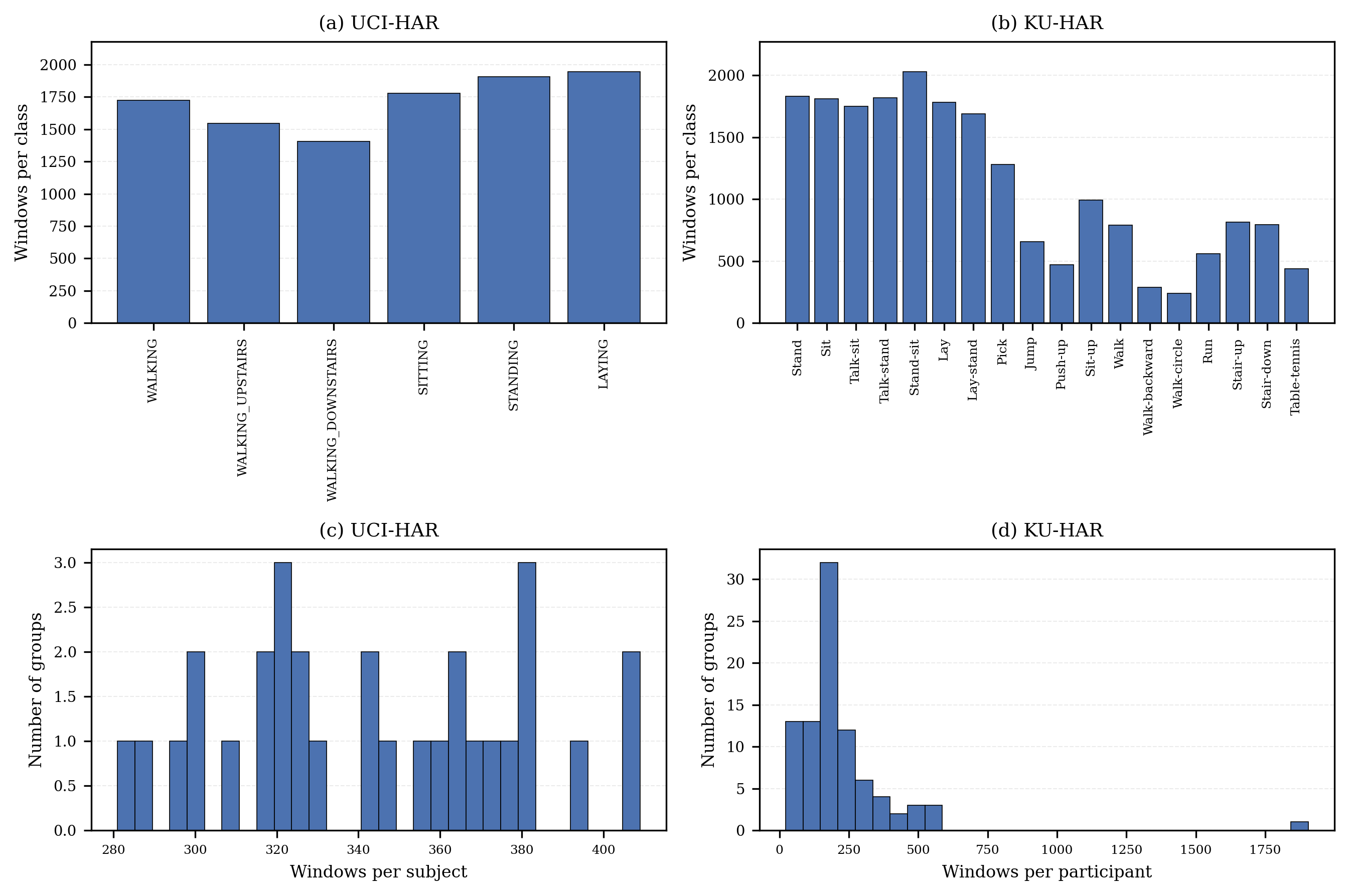}
\caption{Class and group distributions of UCI-HAR (a, c) and KU-HAR (b, d). KU-HAR shows class imbalance and uneven participant contributions.}
\label{fig:datasets}
\end{figure}

The KU-HAR paper reports 90 participants~\cite{Sikder2021KU}; in the public raw release, 89 distinct participant identifiers were found in the filenames. We therefore use 89 participant groups as the grouping unit. After preprocessing, the dataset contains 20{,}037 windows with finite values and valid class labels. The 18 activity classes are: Stand, Sit, Talk-sit, Talk-stand, Stand-sit, Lay, Lay-stand, Pick, Jump, Push-up, Sit-up, Walk, Walk-backward, Walk-circle, Run, Stair-up, Stair-down, and Table-tennis. Figure~\ref{fig:datasets} shows the class and participant distributions of both datasets, and Table~\ref{tab:datasets} summarizes their main characteristics.

\begin{table}[htbp]
\centering
\caption{Dataset characteristics used in the experiments.}
\label{tab:datasets}
\begin{tabular}{lcc}
\hline
\textbf{Characteristic} & \textbf{UCI-HAR} & \textbf{KU-HAR} \\
\hline
Source                & Smartphone inertial sensors & Smartphone inertial sensors \\
Activity classes      & 6 & 18 \\
Sensor channels       & 9 & 6 \\
Sampling frequency    & 50\,Hz & 100\,Hz \\
Window length         & 128 samples & 300 samples \\
Window duration       & 2.56\,s & 3\,s \\
Window overlap        & 50\% & None \\
Number of windows used & 10{,}299 & 20{,}037 \\
Number of groups used & 30 & 89 \\
Main grouping unit    & Subject & Participant \\
\hline
\end{tabular}
\end{table}

\subsection{Data Preprocessing}
The two datasets were loaded and prepared once before model training and stored in a common cached format containing the sensor sequences, activity labels, grouping identifiers, and, for UCI-HAR, the original train/test indicator. No global train/validation/test split or global scaling was performed during this preparation stage. These operations were intentionally deferred to the cross-validation procedure to avoid information leakage. For UCI-HAR, the provided 128-sample windows were directly used as model inputs. For KU-HAR, each raw recording was divided into consecutive non-overlapping 3-s windows. Only complete windows containing 300 samples were retained; any remaining samples shorter than one complete window were discarded. The six sensor channels were stored in the order of three accelerometer channels followed by three gyroscope channels.

For the experiments, three sensor modalities were considered: ACC, GYRO, and ACC+GYRO. ACC and GYRO use their corresponding sensor channels, while ACC+GYRO uses all available inertial channels. For UCI-HAR, ACC comprised the six total-acceleration and body-acceleration channels, GYRO comprised the three body-gyroscope channels, and ACC+GYRO used all nine channels. For KU-HAR, ACC and GYRO each comprised three channels (x, y, z), while ACC+GYRO used all six channels. With 17 features per channel, the resulting feature dimensions were 102, 51, and 153 for ACC, GYRO, and ACC+GYRO on UCI-HAR, and 51, 51, and 102 for the corresponding modalities on KU-HAR. The same modality definitions were applied consistently across the two datasets. No additional signal filtering or handcrafted preprocessing was applied to the raw sequences before the deep learning experiments. Feature extraction for classical machine learning was deterministic and applied to each selected sensor window. The fitted StandardScaler for classical features and channel-wise normalization for deep learning were performed within each training fold, as described in Section~\ref{sec:protocol}.

\subsection{Models and Representations}
\label{sec:models}
\begin{figure}[htbp]
\centering
\begin{subfigure}[b]{0.48\textwidth}
    \centering
    \includegraphics[width=\textwidth]{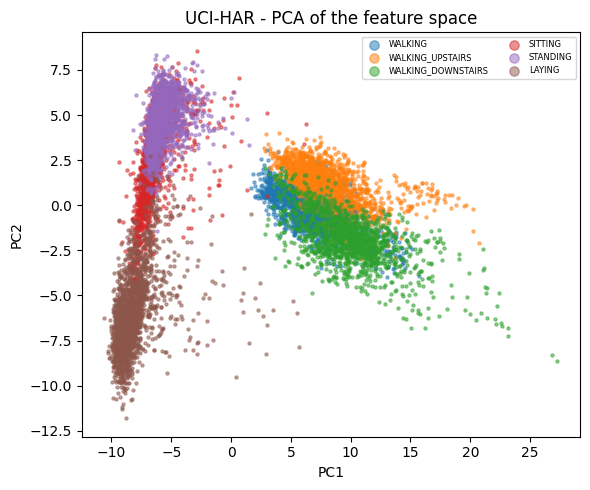}
    \caption{UCI-HAR PCA of feature space}
    \label{fig:rep_uci_pca}
\end{subfigure}
\hfill
\begin{subfigure}[b]{0.48\textwidth}
    \centering
    \includegraphics[width=\textwidth]{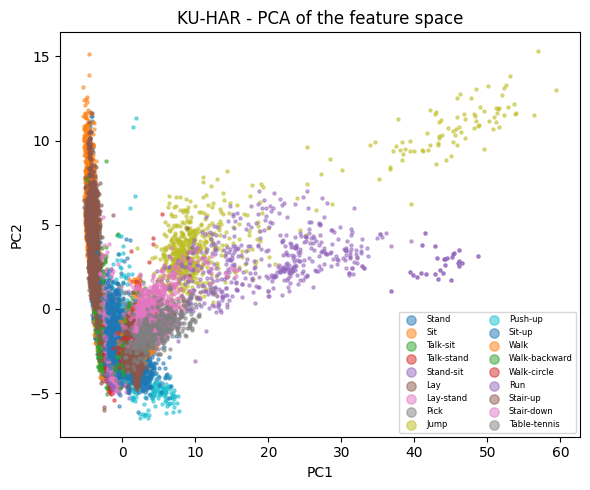}
    \caption{KU-HAR PCA of feature space}
    \label{fig:rep_ku_pca}
\end{subfigure}

\vspace{0.5em}

\begin{subfigure}[b]{\textwidth}
    \centering
    \includegraphics[width=0.85\textwidth]{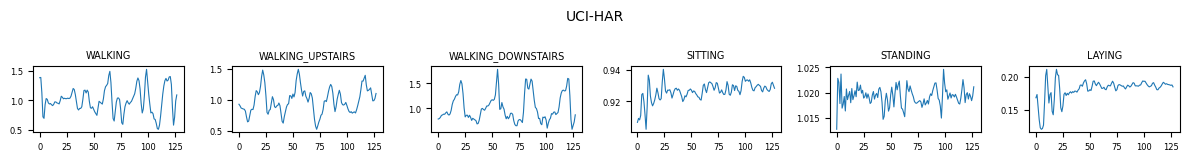}
    \caption{Representative raw sensor sequences from UCI-HAR}
    \label{fig:rep_uci_raw}
\end{subfigure}

\vspace{0.5em}

\begin{subfigure}[b]{\textwidth}
    \centering
    \includegraphics[width=0.85\textwidth]{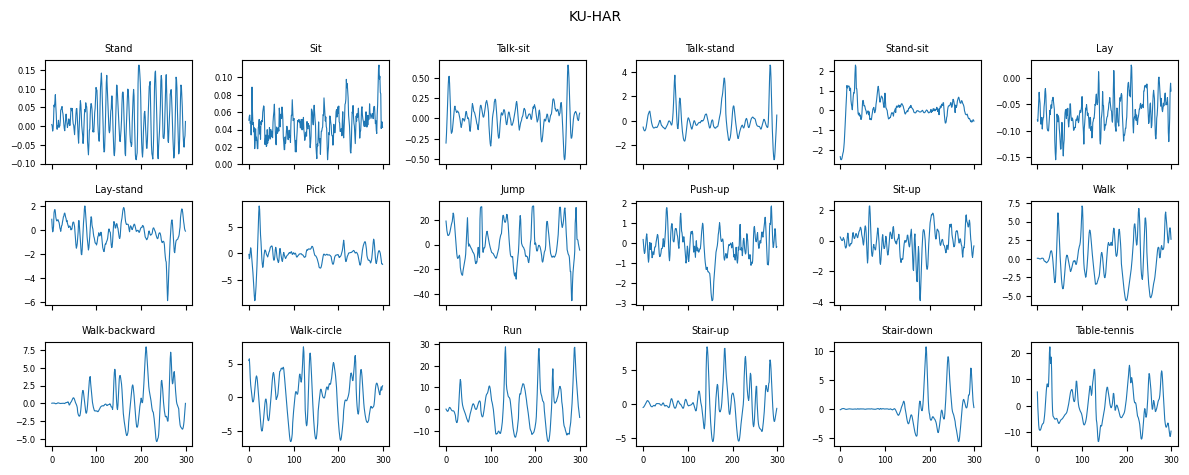}
    \caption{Representative raw sensor sequences from KU-HAR}
    \label{fig:rep_ku_raw}
\end{subfigure}

\caption{Representative visualization of the sensor representations used in the experiments: (a) PCA projection of the UCI-HAR handcrafted feature space, (b) PCA projection of the KU-HAR handcrafted feature space, (c) representative raw sensor sequences from UCI-HAR, and (d) representative raw sensor sequences from KU-HAR.}
\label{fig:representations}
\end{figure}

Two representation strategies were evaluated: handcrafted statistical features for classical machine learning and raw sensor sequences for deep learning. The sensor modalities considered in this study were ACC, GYRO, and their combination ACC+GYRO. Figure~\ref{fig:representations} provides representative visualizations of the two data representations used in this study. The Principal Component Analysis (PCA) projections illustrate the distribution of the handcrafted feature representations, while the example signal plots show the temporal characteristics of the raw sensor sequences provided to the deep learning models.

For the classical machine learning models, each sensor window was converted into a feature vector using 17 handcrafted features per channel. A fixed 17-feature-per-channel recipe was used to maintain a consistent handcrafted feature-extraction procedure across both datasets and the evaluated sensor modalities. Fourteen features were computed in the time domain: mean, standard deviation, minimum, maximum, median, 25th percentile, 75th percentile, interquartile range, root mean square, mean absolute deviation, energy, zero-crossing rate, skewness, and kurtosis. Three additional frequency-domain features were extracted from the power spectrum: dominant frequency, spectral entropy, and spectral centroid. Thus, the resulting feature dimension was $17C$, where $C$ denotes the number of selected sensor channels. Four classifiers were evaluated: Logistic Regression, Random Forest, Support Vector Machine (SVM) with a Radial Basis Function (RBF) kernel, and LightGBM.

\begin{figure}[htbp]
\centering

\begin{minipage}[t]{0.23\textwidth}
\centering
\begin{tikzpicture}[
  block/.style={rectangle, draw, rounded corners, minimum width=2.4cm, minimum height=0.8cm, align=center, font=\scriptsize},
  arr/.style={-{Latex[length=1.5mm]}, thick, draw=gray!80}
]
\node[block] (in) {Input $(C, T)$};
\node[block, below=0.4cm of in] (c1) {Conv1d $C{\to}32$, k5\\BN + ReLU + MaxPool(2)\\Dropout};
\node[block, below=0.4cm of c1] (c2) {Conv1d $32{\to}64$, k5\\BN + ReLU + MaxPool(2)\\Dropout};
\node[block, below=0.4cm of c2] (c3) {Conv1d $64{\to}128$, k3\\BN + ReLU\\AdaptiveAvgPool(1)};
\node[block, below=0.4cm of c3] (fc) {Linear $128{\to}K$};
\draw[arr] (in)--(c1); \draw[arr] (c1)--(c2); \draw[arr] (c2)--(c3); \draw[arr] (c3)--(fc);
\end{tikzpicture}
\caption*{(a) CNN1D}
\end{minipage}
\hfill
\begin{minipage}[t]{0.23\textwidth}
\centering
\begin{tikzpicture}[
  block/.style={rectangle, draw, rounded corners, minimum width=2.4cm, minimum height=0.8cm, align=center, font=\scriptsize},
  arr/.style={-{Latex[length=1.5mm]}, thick, draw=gray!80}
]
\node[block] (in) {Input $(C, T)$};
\node[block, below=0.4cm of in] (perm) {Permute to $(T, C)$};
\node[block, below=0.4cm of perm] (lstm) {2-layer BiLSTM\\$C{\to}64$, bidirectional,\\dropout};
\node[block, below=0.4cm of lstm] (attn) {Attention pooling\\$(T,128){\to}(128)$};
\node[block, below=0.4cm of attn] (fc) {Dropout + Linear\\$128{\to}K$};
\draw[arr] (in)--(perm); \draw[arr] (perm)--(lstm); \draw[arr] (lstm)--(attn); \draw[arr] (attn)--(fc);
\end{tikzpicture}
\caption*{(b) BiLSTM}
\end{minipage}
\hfill
\begin{minipage}[t]{0.23\textwidth}
\centering
\begin{tikzpicture}[
  block/.style={rectangle, draw, rounded corners, minimum width=2.4cm, minimum height=0.75cm, align=center, font=\scriptsize},
  arr/.style={-{Latex[length=1.5mm]}, thick, draw=gray!80}
]
\node[block] (in) {Input $(C, T)$};
\node[block, below=0.35cm of in] (c1) {Conv1d $C{\to}32$, k5\\BN+ReLU+MaxPool(2)};
\node[block, below=0.35cm of c1] (c2) {Conv1d $32{\to}64$, k5\\BN+ReLU+MaxPool(2)};
\node[block, below=0.35cm of c2] (perm) {Permute\\$(T',64)$};
\node[block, below=0.35cm of perm] (lstm) {1-layer BiLSTM\\$64{\to}64$, bidir.};
\node[block, below=0.35cm of lstm] (attn) {Attention\\pooling $\to$(128)};
\node[block, below=0.35cm of attn] (fc) {Dropout+Linear\\$128{\to}K$};
\draw[arr] (in)--(c1); \draw[arr] (c1)--(c2); \draw[arr] (c2)--(perm);
\draw[arr] (perm)--(lstm); \draw[arr] (lstm)--(attn); \draw[arr] (attn)--(fc);
\end{tikzpicture}
\caption*{(c) CNNLSTM}
\end{minipage}
\hfill
\begin{minipage}[t]{0.23\textwidth}
\centering
\begin{tikzpicture}[
  block/.style={rectangle, draw, rounded corners, minimum width=2.4cm, minimum height=0.65cm, align=center, font=\fontsize{6.5pt}{7.5pt}\selectfont},
  arr/.style={-{Latex[length=1.2mm]}, thick, draw=gray!80}
]
\node[block] (in) {Input $(C,T)$};
\node[block, below=0.22cm of in] (c1) {Conv1d $C{\to}64$, k5\\BN+ReLU};
\node[block, below=0.22cm of c1] (c2) {Conv1d $64{\to}64$, k5\\BN+ReLU+MaxPool(2)\\Dropout};
\node[block, below=0.22cm of c2] (c3) {Conv1d $64{\to}128$, k3\\BN+ReLU+MaxPool(2)\\Dropout};
\node[block, below=0.22cm of c3] (perm) {Permute\\$(T',128)$};
\node[block, below=0.22cm of perm] (lstm) {1-layer BiLSTM\\$128{\to}128$, bidir.};
\node[block, below=0.22cm of lstm] (attn) {Attention\\pooling $\to$(256)};
\node[block, below=0.22cm of attn] (fc) {Dropout+Linear\\$256{\to}K$};
\draw[arr] (in)--(c1); \draw[arr] (c1)--(c2); \draw[arr] (c2)--(c3); \draw[arr] (c3)--(perm);
\draw[arr] (perm)--(lstm); \draw[arr] (lstm)--(attn); \draw[arr] (attn)--(fc);
\end{tikzpicture}
\caption*{(d) CNNBiLSTMAttn}
\end{minipage}

\vspace{0.6cm}
\caption{Architectural configurations of the four evaluated deep learning models: (a) CNN1D, (b) BiLSTM, (c) CNNLSTM, and (d) CNNBiLSTMAttn.}
\label{fig:architectures}
\end{figure}

For deep learning, the raw sensor windows were directly provided to four sequence models, as shown in Figure~\ref{fig:architectures}. The input has the form $(C, T)$, where $C$ is the number of sensor channels and $T$ is the number of time steps. Figure~\ref{fig:architectures}(a) presents the CNN1D model, which consists of three one-dimensional convolutional blocks followed by adaptive average pooling and a linear classifier. The model is designed to capture local temporal patterns without recurrent layers. Figure~\ref{fig:architectures}(b) presents the BiLSTM model. The input sequence is first rearranged from $(C, T)$ to $(T, C)$ and processed by a two-layer bidirectional LSTM with a hidden size of 64. The bidirectional outputs are aggregated using learned attention pooling before classification. Figure~\ref{fig:architectures}(c) presents the CNNLSTM model, which combines two convolutional layers with max-pooling and a one-layer bidirectional LSTM. Attention pooling is then used to summarize the temporal representations before the final classifier. Figure~\ref{fig:architectures}(d) presents the CNNBiLSTMAttn model, which uses a deeper convolutional frontend with 64 and 128 channels, followed by a bidirectional LSTM with a hidden size of 128 and attention pooling. This model therefore provides the largest representational capacity among the evaluated deep architectures. All deep learning models use dropout and a final linear layer whose output dimension corresponds to the number of activity classes. The training configuration is described in Section~\ref{sec:training}.

For the primary model comparison, all eight models were evaluated using the common ACC+GYRO modality. For modality analysis, ACC, GYRO, and ACC+GYRO were evaluated for the four classical machine learning models and CNNBiLSTMAttn, while CNN1D, BiLSTM, and CNNLSTM were evaluated only with ACC+GYRO. CNNBiLSTMAttn was used as the representative deep learning architecture for the modality analysis.

\subsubsection{Training Configuration}
\label{sec:training}
All deep learning models were implemented in PyTorch~2.7.1 and trained on a single NVIDIA GeForce RTX~3050 Laptop GPU with 4\,GB of VRAM. The models were optimized with AdamW with a learning rate of $1\times10^{-3}$ and a batch size of 64, for a maximum of 100 epochs. Cross-entropy loss was used for multi-class classification. The dropout rates were fixed at 0.3 for CNN1D, BiLSTM and CNNLSTM, and 0.4 for CNNBiLSTMAttn. Early stopping with a patience of 10 epochs was applied based on validation macro-F1, and the checkpoint with the highest validation macro-F1 was retained for evaluation on the outer-test fold (Section~\ref{sec:protocol}). The same training configuration was applied consistently to all evaluated deep learning configurations. Classical machine learning models were implemented using scikit-learn~1.7.0, and LightGBM used version~4.7.0 of its scikit-learn-compatible classifier. Logistic Regression was configured with a maximum of 1{,}000 iterations, Random Forest with 300 trees, SVM with an RBF kernel ($C=10$, $\gamma=\mathrm{scale}$), and LightGBM with 300 estimators. No hyperparameter tuning was performed for any model. The base random seed was 42. For each cross-validation fold, a derived seed was computed as $100 \times 42 + \text{fold index}$, ensuring deterministic and reproducible per-fold initialization. All random number generators (Python, NumPy, and PyTorch) were seeded accordingly, and cuDNN was configured for deterministic execution.

\subsection{Experimental Protocol}
\label{sec:protocol}
\begin{figure}[htbp]
\centering
\resizebox{\textwidth}{!}{%
\begin{tikzpicture}[
    node distance=0.55cm and 1.0cm,
    block/.style={
        rectangle, rounded corners=3pt,
        draw=black!70, thick,
        minimum width=7.2cm, minimum height=0.72cm,
        align=center, font=\sffamily\small
    },
    split/.style={
        rectangle, rounded corners=3pt,
        draw=black!60, thick,
        minimum width=3.2cm, minimum height=0.65cm,
        align=center, font=\sffamily\small
    },
    role/.style={
        rectangle, rounded corners=3pt,
        draw=black!70, thick,
        minimum width=3.5cm, minimum height=0.85cm,
        align=center, font=\sffamily\small
    },
    note/.style={
        rectangle, rounded corners=3pt,
        draw=black!50, dashed,
        minimum width=7.2cm, minimum height=0.6cm,
        align=center, font=\sffamily\footnotesize
    },
    arr/.style={-{Latex[length=2mm]}, thick, draw=black!70}
]
\node[block] (A) {Full dataset $(X, y, \mathrm{groups})$};
\node[block, below=of A] (B)
    {Outer 5-fold StratifiedGroupKFold\\
     Groups: subject (UCI-HAR) / participant (KU-HAR)};
\node[split, below left=0.5cm and 0.8cm of B] (C1)
    {Outer-training\\(4 folds)};
\node[split, below right=0.5cm and 0.8cm of B] (C2)
    {Outer-test\\(1 held-out fold)};
\node[block, below=0.6cm of C1, minimum width=6.4cm] (D)
    {Inner grouped split ($\sim$80/20)};
\node[split, below left=0.5cm and 0.3cm of D] (E1) {Inner-training};
\node[split, below right=0.5cm and 0.3cm of D] (E2) {Inner-validation};
\node[role, below=0.7cm of E1] (F)
    {Fit preprocessing and model\\on inner-training only};
\node[role, below=0.7cm of E2] (V)
    {Validation / diagnostic\\using inner-validation};
\path (F) -- (V) coordinate[midway] (FVmid);
\node[block, below=1.0cm of FVmid] (G)
    {Deep learning: early stopping / checkpoint selection\\
     Classical ML: single fit; inner-validation used only for diagnostics};
\node[block, below=0.7cm of G] (I)
    {Evaluate once on outer-test fold};
\node[note, below=0.35cm of I] (J)
    {No tuning on outer-test; all choices fixed beforehand};
\draw[arr] (A) -- (B);
\draw[arr] (B) -| (C1);
\draw[arr] (B) -| (C2);
\draw[arr] (C1) -- (D);
\draw[arr] (D) -| (E1);
\draw[arr] (D) -| (E2);
\draw[arr] (E1) -- (F);
\draw[arr] (E2) -- (V);
\draw[arr] (F) |- (G.west);
\draw[arr] (V) |- (G.east);
\draw[arr] (G) -- (I);
\draw[arr] (C2) |- (I);
\draw[arr] (I) -- (J);
\end{tikzpicture}%
}
\caption{Group-aware experimental evaluation protocol used for the primary comparison.}
\label{fig:protocol}
\end{figure}
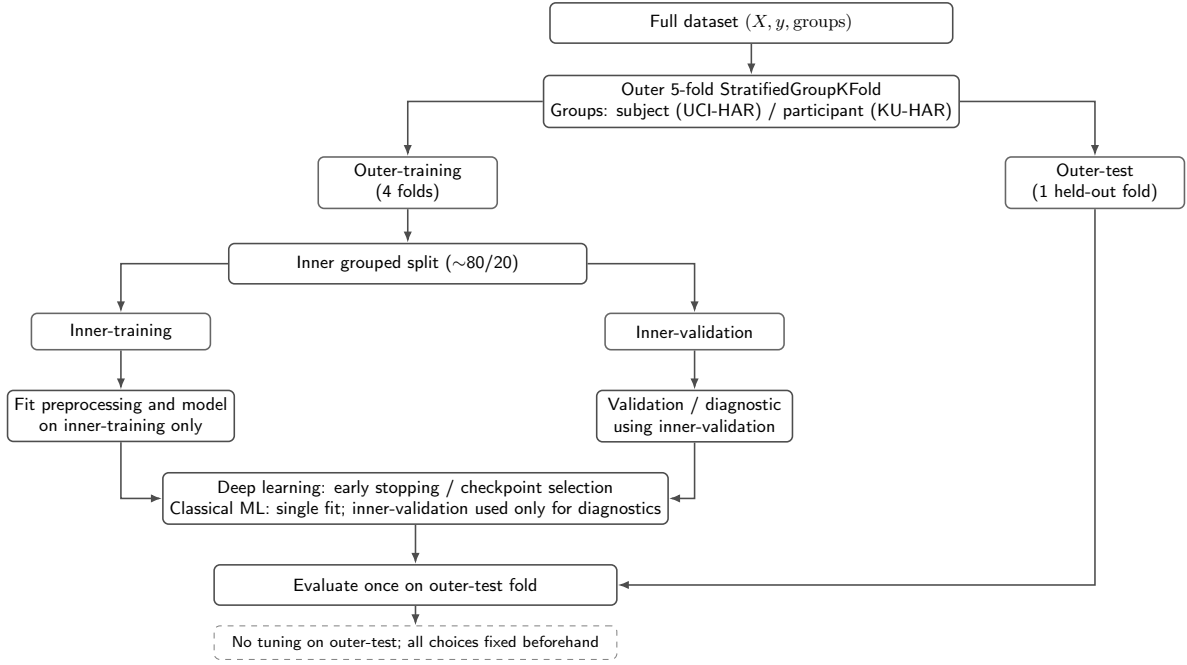

The primary evaluation used five-fold StratifiedGroupKFold cross-validation to assess model performance under group-independent evaluation, as summarized in Figure~\ref{fig:protocol}. For UCI-HAR, subjects were used as the grouping unit, while participant identifiers were used for KU-HAR. Thus, samples from the same subject or participant were not shared between training and outer-test folds. The cross-validation procedure was implemented using StratifiedGroupKFold with five folds. For each outer fold, four of the five folds were used for model development (outer-training) and the remaining fold was held out for evaluation (outer-test). The outer-training data were further divided into inner-training and inner-validation sets using a grouped split of $\sim$80/20. All preprocessing parameters that require fitting were obtained from the inner-training data only and then applied unchanged to the corresponding validation and outer-test data. This procedure prevents information from the held-out test fold from influencing preprocessing or model development.

For deep learning models, early stopping and checkpoint selection followed the configuration described in Section~\ref{sec:training}. Classical machine learning models were fitted using the inner-training data, while the inner-validation set was used only for diagnostic evaluation. The final model performance for every fold was always measured on the outer-test fold, not on the inner-validation set. All architectural and training choices were fixed before evaluation. For classical machine learning, feature vectors were standardized using a StandardScaler fitted on the inner-training data and then applied unchanged to the validation and outer-test data. For deep learning, each selected sensor channel was standardized using its mean and standard deviation computed from the inner-training samples and then applied unchanged to the validation and outer-test data. No hyperparameter tuning was performed using the outer-test data. The primary results therefore represent the performance of the predefined model configurations under the same group-aware evaluation procedure.

Two additional protocols were evaluated for methodological context. First, the official train/test partition of UCI-HAR was used as a secondary evaluation to facilitate comparison with studies using the original dataset split. Second, a random stratified split that ignored group identities was evaluated as a leakage ablation. This secondary experiment used a single representative configuration (CNNBiLSTMAttn with ACC+GYRO) on the same data and preprocessing as the grouped evaluation, and was intended to quantify the effect of ignoring subject/participant boundaries.

\subsection{Evaluation Metrics}
Model performance was evaluated using accuracy, macro-precision, macro-recall, and macro-F1. For a multi-class classification problem, let $TP_c$, $FP_c$, and $FN_c$ denote the true positives, false positives, and false negatives for class $c$. Precision and recall for each class are defined as
\begin{equation}
\mathrm{Precision}_c = \frac{TP_c}{TP_c + FP_c},
\end{equation}
\begin{equation}
\mathrm{Recall}_c = \frac{TP_c}{TP_c + FN_c}.
\end{equation}
The F1-score for class $c$ is
\begin{equation}
F1_c = \frac{2 \cdot \mathrm{Precision}_c \cdot \mathrm{Recall}_c}{\mathrm{Precision}_c + \mathrm{Recall}_c}.
\end{equation}
For $K$ activity classes, macro-averaged precision, recall, and F1 are calculated by giving equal weight to every class:
\begin{equation}
\mathrm{Macro\text{-}Precision} = \frac{1}{K} \sum_{c=1}^{K} \mathrm{Precision}_c,
\end{equation}
\begin{equation}
\mathrm{Macro\text{-}Recall} = \frac{1}{K} \sum_{c=1}^{K} \mathrm{Recall}_c,
\end{equation}
\begin{equation}
\mathrm{Macro\text{-}F1} = \frac{1}{K} \sum_{c=1}^{K} F1_c.
\end{equation}
Accuracy is defined as the proportion of correctly classified samples:
\begin{equation}
\mathrm{Accuracy} = \frac{N_{\mathrm{correct}}}{N_{\mathrm{total}}}.
\end{equation}
Macro-F1 was selected as the primary evaluation metric because it gives equal importance to all activity classes and is therefore less dominated by frequent classes than accuracy. All reported metrics were computed on the outer-test fold in each cross-validation iteration, and the final cross-validation results were summarized as the mean and standard deviation across the five outer folds.

\section{Results}

\subsection{Overall Model Comparison}
Table~\ref{tab:overall} summarizes the macro-F1 performance of all evaluated models under the ACC+GYRO modality using the primary grouped cross-validation protocol. This modality provides a common setting for comparing the four classical machine learning models and the four deep learning models across both datasets. Each result is reported as the mean and standard deviation over five outer folds.

On UCI-HAR, all models achieved relatively high macro-F1 values, ranging from 0.910 for Random Forest to 0.936 for CNNLSTM. The deep learning models produced values between 0.932 and 0.936, while the classical machine learning models ranged from 0.910 to 0.929. LightGBM achieved $0.929 \pm 0.021$, which was close to the deep learning results. The four deep learning models also showed relatively small differences in their mean performance, with CNNLSTM obtaining $0.936 \pm 0.031$, CNNBiLSTMAttn $0.935 \pm 0.034$, CNN1D $0.934 \pm 0.041$, and BiLSTM $0.932 \pm 0.027$. In contrast, KU-HAR showed lower and more variable performance across models. The classical machine learning models achieved macro-F1 values from 0.810 to 0.857, while the deep learning models ranged from 0.777 to 0.828. LightGBM obtained $0.857 \pm 0.131$, followed by SVM at $0.847 \pm 0.112$ and Random Forest at $0.836 \pm 0.119$. Among the deep learning models, CNNLSTM obtained $0.828 \pm 0.112$, while CNN1D achieved $0.777 \pm 0.142$. These results show that the relative performance of classical and deep learning models was dataset-dependent. Deep learning models performed strongly on UCI-HAR, but they did not consistently outperform the classical feature-based models on KU-HAR. Although this difference is within the standard deviation of both configurations, the grouped protocol used here ensures that no participant is shared between training and test folds, making the comparison indicative of group-independent performance. The results therefore indicate that model family alone does not determine performance, motivating further analysis of sensor modality and dataset characteristics in the following sections.

\begin{table}[htbp]
\centering
\caption{Overall macro-F1 performance under the ACC+GYRO modality. Values are mean $\pm$ standard deviation over five grouped cross-validation folds.}
\label{tab:overall}
\begin{tabular}{lcc}
\hline
\textbf{Model} & \textbf{UCI-HAR} & \textbf{KU-HAR} \\
\hline
Logistic Regression & $0.925 \pm 0.031$ & $0.810 \pm 0.127$ \\
Random Forest       & $0.910 \pm 0.021$ & $0.836 \pm 0.119$ \\
SVM                 & $0.922 \pm 0.027$ & $0.847 \pm 0.112$ \\
LightGBM            & $0.929 \pm 0.021$ & $0.857 \pm 0.131$ \\
CNN1D               & $0.934 \pm 0.041$ & $0.777 \pm 0.142$ \\
BiLSTM              & $0.932 \pm 0.027$ & $0.788 \pm 0.118$ \\
CNNLSTM             & $0.936 \pm 0.031$ & $0.828 \pm 0.112$ \\
CNNBiLSTMAttn       & $0.935 \pm 0.034$ & $0.801 \pm 0.107$ \\
\hline
\end{tabular}
\end{table}

\subsection{Sensor Modality Comparison}
\begin{figure}[!ht]
\centering
\begin{subfigure}[b]{\textwidth}
    \centering
    \includegraphics[width=0.85\textwidth]{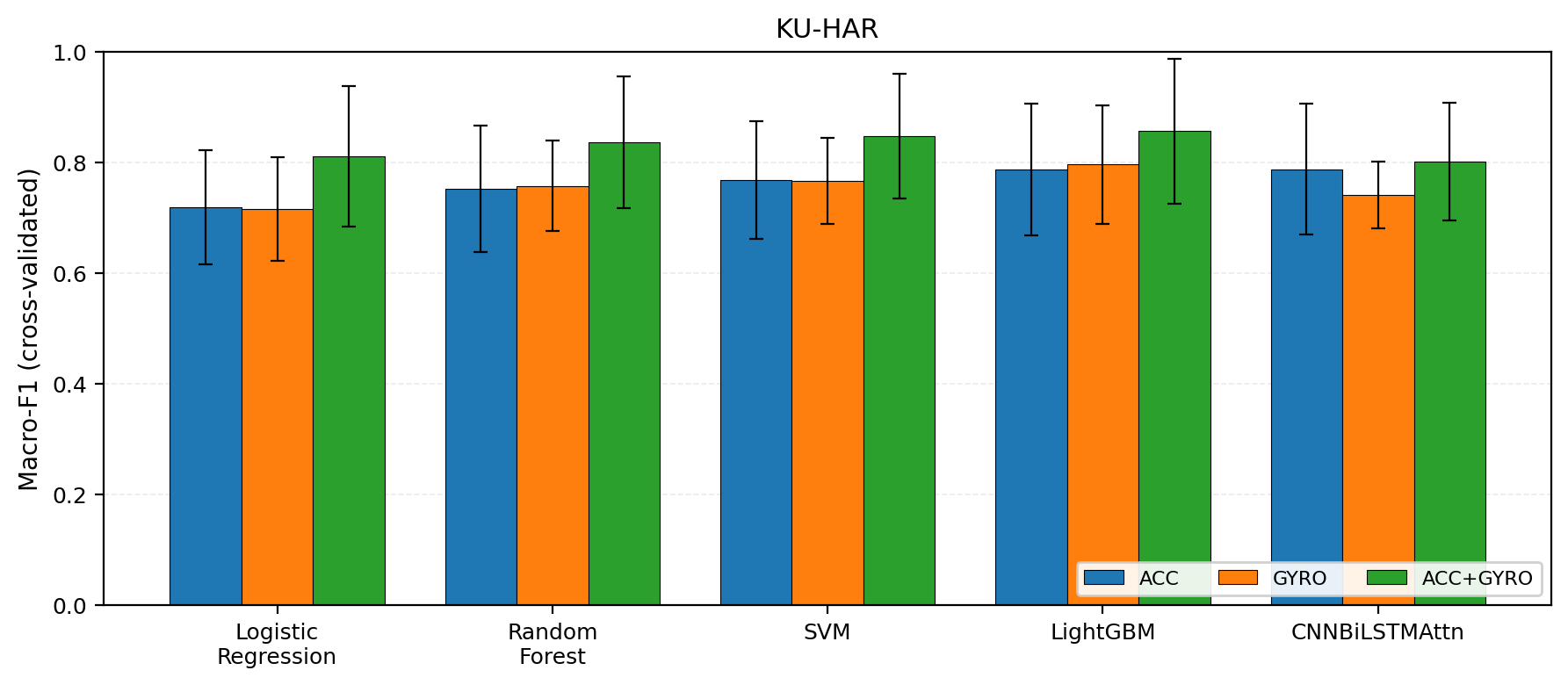}
    \caption{KU-HAR}
    \label{fig:modality_ku}
\end{subfigure}

\vspace{0.5em}

\begin{subfigure}[b]{\textwidth}
    \centering
    \includegraphics[width=0.85\textwidth]{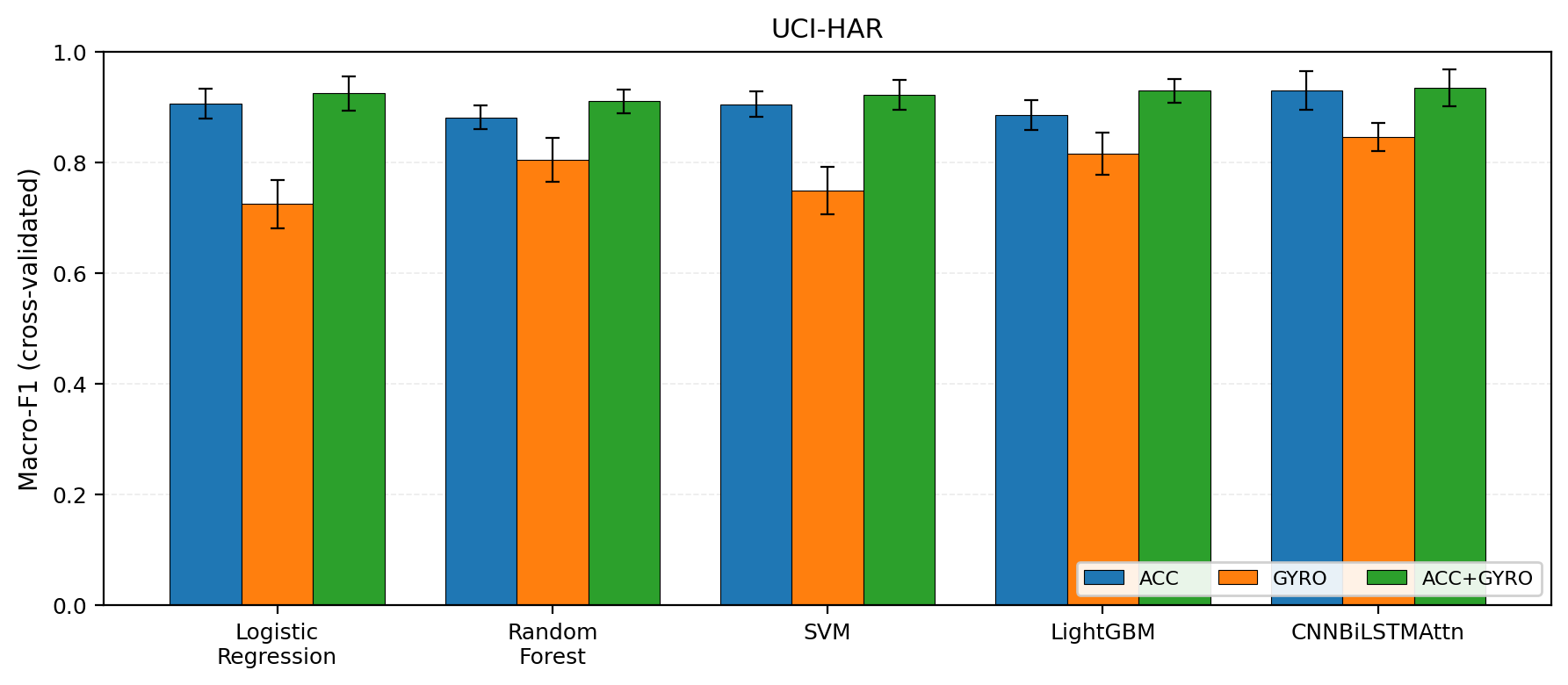}
    \caption{UCI-HAR}
    \label{fig:modality_uci}
\end{subfigure}

\caption{Macro-F1 performance across sensor modalities under grouped five-fold cross-validation. Bars show mean macro-F1 and error bars indicate standard deviation across the five outer folds.}
\label{fig:modality}
\end{figure}

Figure~\ref{fig:modality} and Table~\ref{tab:modality} compare the effect of sensor modality using grouped five-fold cross-validation. Figure~\ref{fig:modality} visualizes the mean macro-F1 and fold-level variability across the evaluated models and modalities, while Table~\ref{tab:modality} provides the corresponding numerical results. The analysis focuses on the five models that were evaluated using ACC, GYRO, and ACC+GYRO: four classical machine learning models and CNNBiLSTMAttn. Across both datasets, combining accelerometer and gyroscope signals generally produced higher macro-F1 than using either sensor alone. On KU-HAR, ACC+GYRO improved the macro-F1 of Logistic Regression from 0.719 to 0.810, Random Forest from 0.753 to 0.836, SVM from 0.768 to 0.847, LightGBM from 0.787 to 0.857, and CNNBiLSTMAttn from 0.788 to 0.801. A similar pattern was observed on UCI-HAR. The combined modality increased macro-F1 relative to ACC for all five models, although the improvement varied in magnitude. Compared with GYRO alone, the gain was substantially larger for the classical machine learning models, particularly Logistic Regression and SVM, whereas CNNBiLSTMAttn showed the smallest improvement among the five evaluated configurations. The combined modality achieved 0.925 for Logistic Regression, 0.910 for Random Forest, 0.922 for SVM, 0.929 for LightGBM, and 0.935 for CNNBiLSTMAttn.

\begin{table}[htbp]
\centering
\caption{Macro-F1 performance across sensor modalities under grouped five-fold cross-validation. Values are mean $\pm$ standard deviation.}
\label{tab:modality}
\small
\begin{tabular}{llccc}
\hline
\textbf{Dataset} & \textbf{Model} & \textbf{ACC} & \textbf{GYRO} & \textbf{ACC+GYRO} \\
\hline
KU-HAR & Logistic Regression & $0.719 \pm 0.104$ & $0.716 \pm 0.094$ & $0.810 \pm 0.127$ \\
       & Random Forest       & $0.753 \pm 0.115$ & $0.757 \pm 0.082$ & $0.836 \pm 0.119$ \\
       & SVM                 & $0.768 \pm 0.106$ & $0.767 \pm 0.077$ & $0.847 \pm 0.112$ \\
       & LightGBM            & $0.787 \pm 0.119$ & $0.796 \pm 0.108$ & $0.857 \pm 0.131$ \\
       & CNNBiLSTMAttn       & $0.788 \pm 0.118$ & $0.741 \pm 0.060$ & $0.801 \pm 0.107$ \\
\hline
UCI-HAR & Logistic Regression & $0.906 \pm 0.027$ & $0.724 \pm 0.044$ & $0.925 \pm 0.031$ \\
        & Random Forest       & $0.881 \pm 0.022$ & $0.805 \pm 0.040$ & $0.910 \pm 0.021$ \\
        & SVM                 & $0.905 \pm 0.023$ & $0.749 \pm 0.042$ & $0.922 \pm 0.027$ \\
        & LightGBM            & $0.886 \pm 0.026$ & $0.816 \pm 0.038$ & $0.929 \pm 0.021$ \\
        & CNNBiLSTMAttn       & $0.930 \pm 0.035$ & $0.846 \pm 0.025$ & $0.935 \pm 0.034$ \\
\hline
\end{tabular}
\end{table}

The results also indicate that the contribution of individual sensor modalities is dataset- and model-dependent. For example, on KU-HAR, LightGBM performed slightly better with GYRO (0.796) than with ACC (0.787), whereas CNNBiLSTMAttn showed the opposite pattern, with 0.788 using ACC and 0.741 using GYRO. On UCI-HAR, ACC consistently produced higher macro-F1 than GYRO for the models included in this modality analysis. Overall, the results suggest that complementary information from accelerometer and gyroscope signals is beneficial under the evaluated grouped protocol, while the relative usefulness of an individual modality depends on the dataset and model.

\subsection{Grouped vs. Random Split}
\begin{figure}[htbp]
\centering
\begin{subfigure}[b]{0.48\textwidth}
    \centering
    \includegraphics[width=\textwidth]{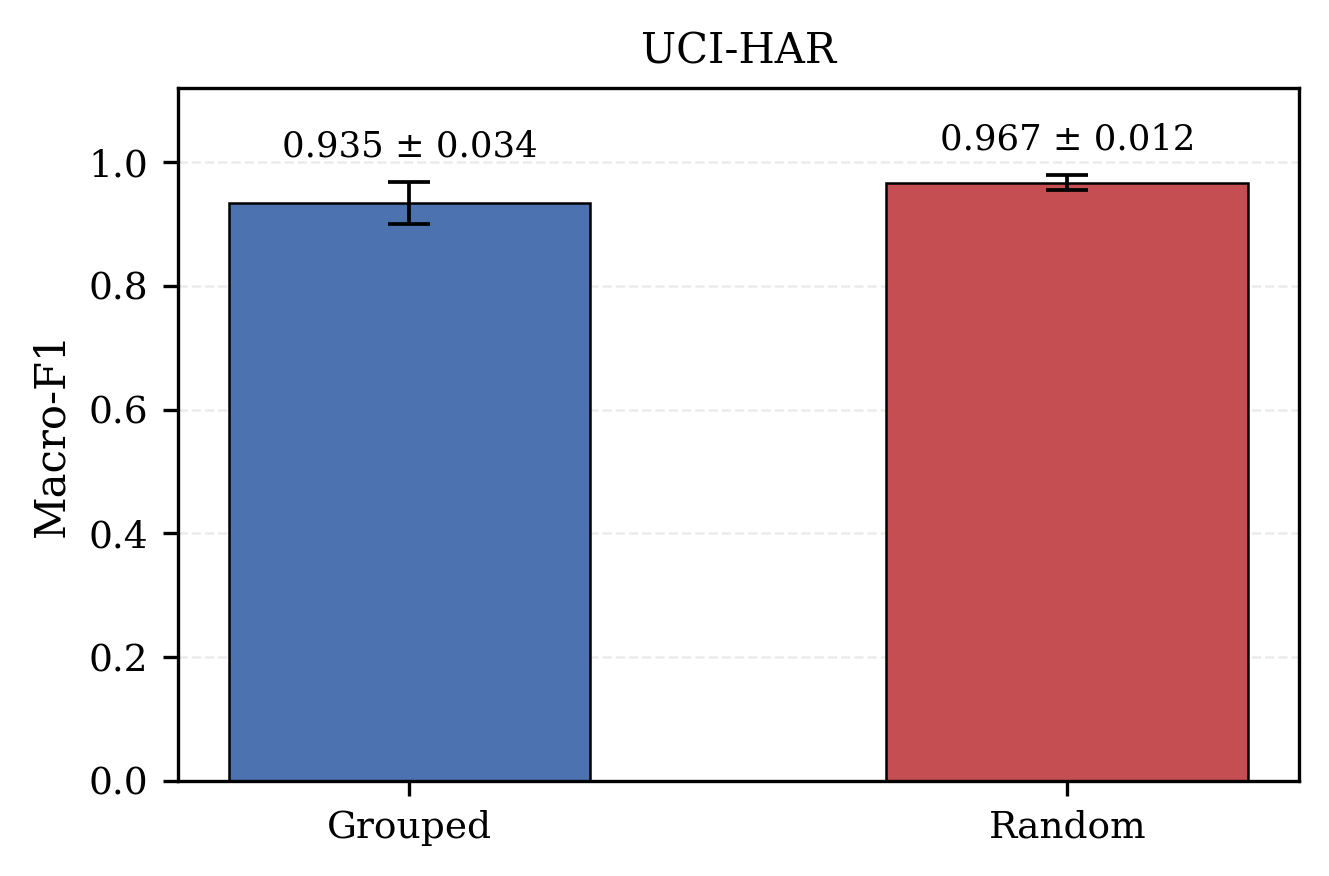}
    \caption{UCI-HAR}
    \label{fig:leakage_uci}
\end{subfigure}
\hfill
\begin{subfigure}[b]{0.48\textwidth}
    \centering
    \includegraphics[width=\textwidth]{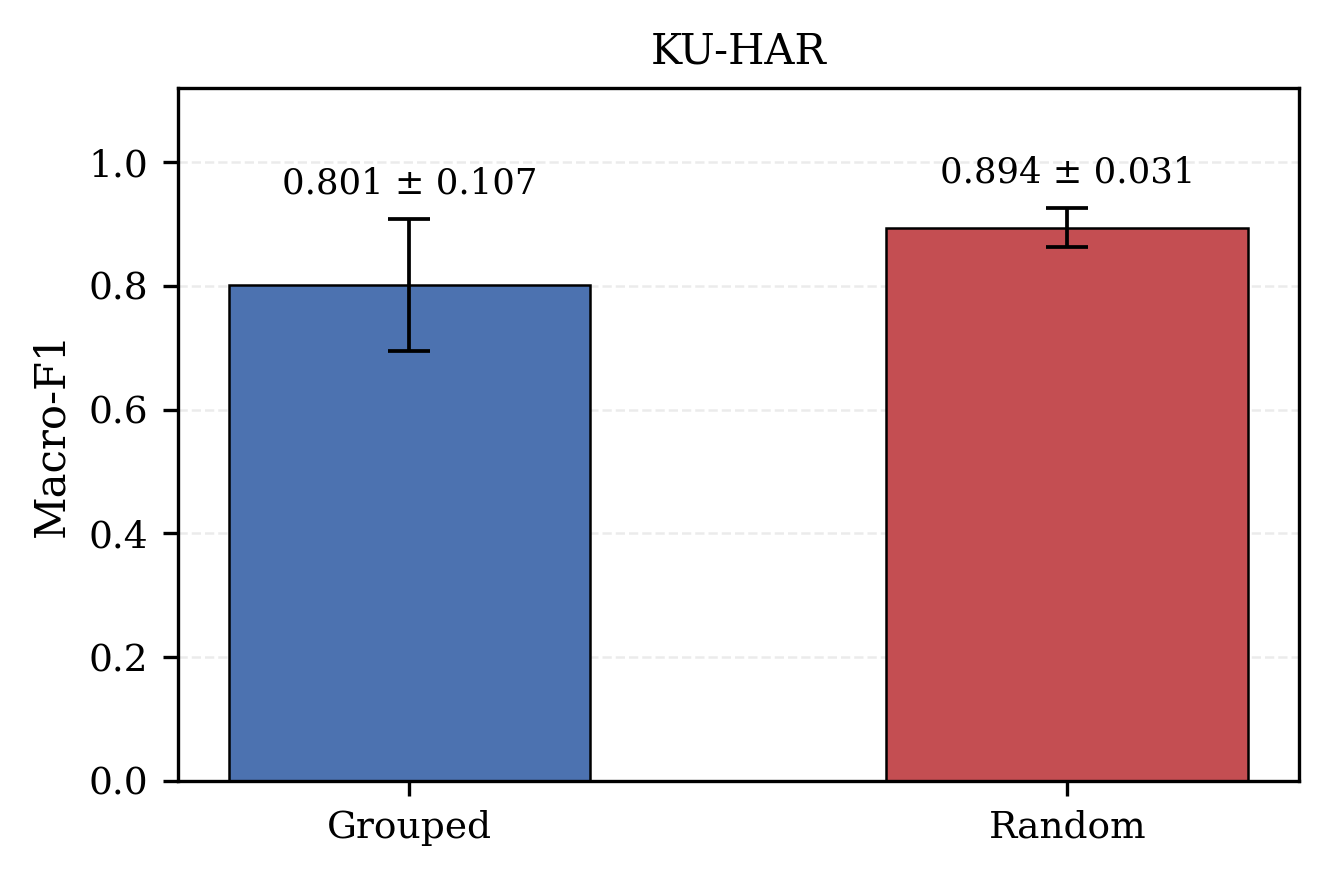}
    \caption{KU-HAR}
    \label{fig:leakage_ku}
\end{subfigure}
\caption{Comparison of grouped and random split evaluation for CNNBiLSTMAttn with ACC+GYRO on (a) UCI-HAR and (b) KU-HAR. Values are mean macro-F1 $\pm$ standard deviation over five folds.}
\label{fig:leakage}
\end{figure}

The effect of the evaluation protocol was examined by comparing grouped and random stratified splits using the same model configuration and sensor modality. This analysis used CNNBiLSTMAttn with ACC+GYRO, which was evaluated under both protocols. As reported in Table~\ref{tab:leakage} and illustrated in Figure~\ref{fig:leakage}, the random split produced higher macro-F1 values than the grouped split for both datasets. On UCI-HAR, macro-F1 increased from $0.935 \pm 0.034$ under grouped evaluation to $0.967 \pm 0.012$ under random evaluation, corresponding to an increase of 0.033. On KU-HAR, macro-F1 increased from $0.801 \pm 0.107$ to $0.894 \pm 0.031$, corresponding to an increase of 0.093. The difference was therefore larger on KU-HAR than on UCI-HAR. Under the random protocol, group identities are ignored, allowing samples from the same subject or participant to be distributed across training and test folds. This produces a less strict evaluation of performance on unseen groups. In contrast, the grouped protocol preserves subject or participant boundaries between training and test folds. The lower macro-F1 obtained under this protocol therefore provides a stricter measure of group-independent performance. The grouped protocol was consequently retained as the primary evaluation setting, while the random split was used only as a methodological ablation. This analysis is limited to CNNBiLSTMAttn with ACC+GYRO and should therefore be interpreted as an assessment of the effect of the split protocol for this configuration rather than as a universal estimate across all evaluated models.

\begin{table}[htbp]
\centering
\caption{Macro-F1 comparison between grouped and random split evaluation for CNNBiLSTMAttn with ACC+GYRO. Values are mean $\pm$ standard deviation over five folds.}
\label{tab:leakage}
\begin{tabular}{lccc}
\hline
\textbf{Dataset} & \textbf{Grouped} & \textbf{Random} & \textbf{Increase} \\
\hline
UCI-HAR & $0.935 \pm 0.034$ & $0.967 \pm 0.012$ & 0.033 \\
KU-HAR  & $0.801 \pm 0.107$ & $0.894 \pm 0.031$ & 0.093 \\
\hline
\end{tabular}
\end{table}

\subsection{Error Analysis}
\begin{figure}[p]
\centering
\begin{subfigure}[b]{\textwidth}
    \centering
    \includegraphics[width=0.9\textwidth]{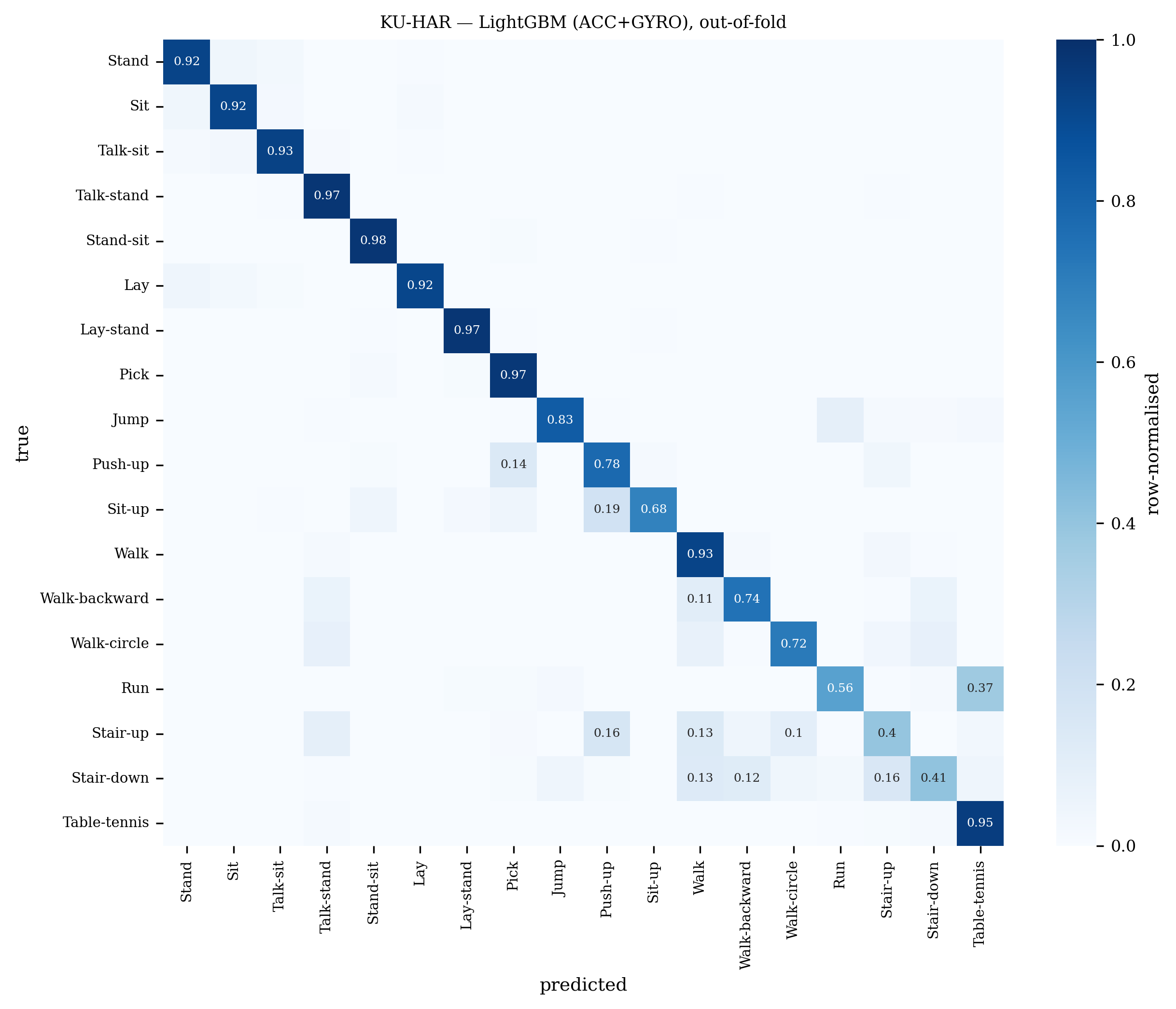}
    \caption{KU-HAR (LightGBM, ACC+GYRO)}
    \label{fig:confusion_ku}
\end{subfigure}

\vspace{0.8em}

\begin{subfigure}[b]{\textwidth}
    \centering
    \includegraphics[width=0.7\textwidth]{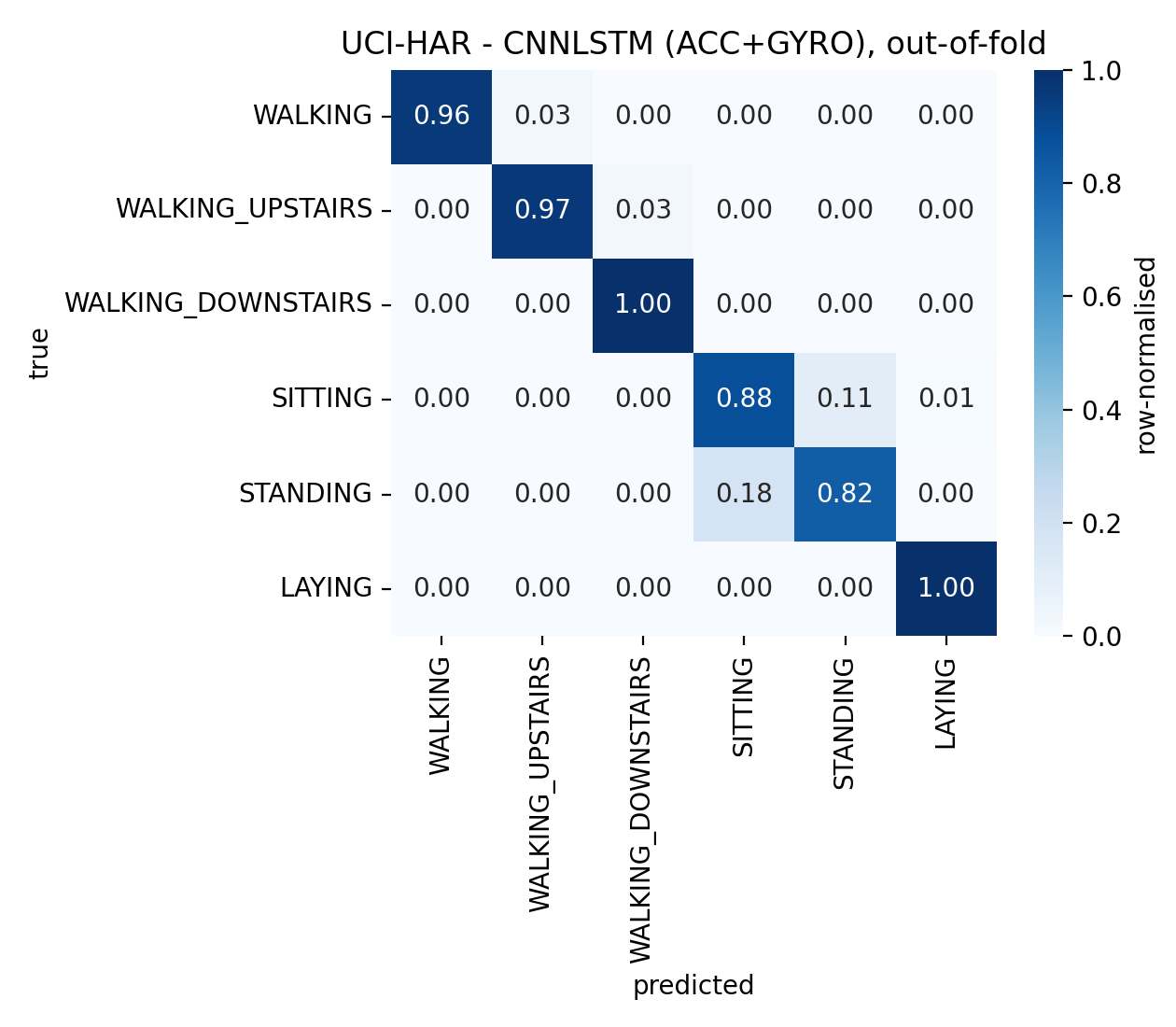}
    \caption{UCI-HAR (CNNLSTM, ACC+GYRO)}
    \label{fig:confusion_uci}
\end{subfigure}

\caption{Row-normalized confusion matrices from pooled out-of-fold predictions for the best-performing configuration on (a) KU-HAR (LightGBM, ACC+GYRO) and (b) UCI-HAR (CNNLSTM, ACC+GYRO).}
\label{fig:confusion}
\end{figure}

Error patterns were further examined using row-normalized confusion matrices constructed from pooled out-of-fold predictions. As shown in Figure~\ref{fig:confusion}, the analysis provides the aggregated prediction behavior of each sample when it served as the outer-test sample, rather than relying on a single test split. For KU-HAR, the confusion matrix of LightGBM with ACC+GYRO shows strong diagonal performance for many activities, particularly relatively distinct activities such as Stand, Sit, Talk-sit, Talk-stand, Stand-sit, Lay-stand, and Table-tennis. However, several dynamic activities remain more difficult to distinguish. The lowest class recalls were observed for Stair-up (0.398), Stair-down (0.407), and Run (0.555), indicating substantial confusion among these movement-related activities. For UCI-HAR, CNNLSTM with ACC+GYRO produced a stronger and more concentrated diagonal pattern. WALKING, WALKING\_UPSTAIRS, and WALKING\_DOWNSTAIRS were classified with high recall, while the main errors occurred among the static activities. In particular, STANDING had the lowest recall at approximately 0.822, followed by SITTING at approximately 0.881. The confusion matrix shows that these two activities were frequently confused with each other. The error patterns therefore differ between datasets. KU-HAR exhibits more difficulty among several fine-grained dynamic activities, whereas UCI-HAR shows a more limited set of errors, primarily involving similar static postures.

\clearpage

\clearpage

\subsection{Computational Cost}
Table~\ref{tab:cost} summarizes the computational characteristics of the evaluated deep learning models in terms of trainable parameters, mean fitting time, and the highest mean macro-F1 among the evaluated configurations. The parameter counts reflect model size, while fitting time provides a practical measure of training cost under the experimental setup. CNN1D had the smallest number of parameters on both datasets (38{,}770 on KU-HAR, 37{,}702 on UCI-HAR) and the shortest mean fitting time on KU-HAR (44.3\,s). CNNBiLSTMAttn had the largest model size (316{,}179 parameters on KU-HAR and 313{,}735 on UCI-HAR) and required longer fitting time than CNN1D and CNNLSTM. CNNLSTM provided a relatively moderate computational cost while achieving the highest observed macro-F1 among the deep learning models on both datasets: 0.8284 on KU-HAR and 0.9359 on UCI-HAR. On KU-HAR, its mean fitting time was 89.8\,s with 80{,}499 parameters, while on UCI-HAR it required 17.7\,s with 79{,}431 parameters. The fitting times were substantially higher on KU-HAR than on UCI-HAR for all deep learning models. This difference reflects the larger number of windows (20{,}037 vs 10{,}299) and the longer input sequences (300 vs 128 samples) used on KU-HAR. Computational cost should therefore be considered together with predictive performance rather than evaluated independently.

\begin{table}[htbp]
\centering
\caption{Computational cost and best mean macro-F1 of the evaluated deep learning models. Parameter counts are fixed model properties. Mean fitting time is averaged over five folds; for CNNBiLSTMAttn, it is averaged across its three evaluated modalities, while the remaining models are evaluated using ACC+GYRO only. ``Best mean Macro-F1'' denotes the highest mean macro-F1 among the evaluated modality configurations.}
\label{tab:cost}
\small
\begin{tabular}{llrrr}
\hline
\textbf{Dataset} & \textbf{Model} & \textbf{Parameters} & \textbf{Mean Fit Time (s)} & \textbf{Best Macro-F1} \\
\hline
KU-HAR & BiLSTM          & 138{,}643 & 132.3 & 0.7881 \\
       & CNN1D           &  38{,}770 &  44.3 & 0.7769 \\
       & CNNBiLSTMAttn   & 316{,}179 & 107.8 & 0.8014 \\
       & CNNLSTM         &  80{,}499 &  89.8 & 0.8284 \\
\hline
UCI-HAR & BiLSTM         & 138{,}631 &  12.7 & 0.9325 \\
        & CNN1D          &  37{,}702 &  15.6 & 0.9342 \\
        & CNNBiLSTMAttn  & 313{,}735 &  33.0 & 0.9346 \\
        & CNNLSTM        &  79{,}431 &  17.7 & 0.9359 \\
\hline
\end{tabular}
\end{table}

\subsection{Secondary Evaluation on the Official UCI-HAR Split}
As a secondary analysis, the original UCI-HAR train/test partition was evaluated using the same predefined model configurations. Table~\ref{tab:official} reports the resulting macro-F1 values. CNN1D obtained the highest macro-F1 (0.926), followed by LightGBM and SVM (0.924), while the other deep learning models ranged from 0.899 to 0.917. Because this analysis uses a single fixed partition rather than five-fold cross-validation, the results are presented without fold-level variability and are not used for the primary model comparison. The primary conclusions therefore remain based on the grouped cross-validation results.

\begin{table}[htbp]
\centering
\caption{Secondary evaluation using the official UCI-HAR train/test partition.}
\label{tab:official}
\begin{tabular}{lc}
\hline
\textbf{Model} & \textbf{Macro-F1} \\
\hline
Logistic Regression & 0.920 \\
Random Forest       & 0.899 \\
SVM                 & 0.924 \\
LightGBM            & 0.924 \\
CNN1D               & 0.926 \\
BiLSTM              & 0.899 \\
CNNLSTM             & 0.900 \\
CNNBiLSTMAttn       & 0.917 \\
\hline
\end{tabular}
\end{table}

\noindent\textit{Note: The official split was evaluated as a secondary analysis. The reported values are from a single split and therefore are not accompanied by cross-validation standard deviations. An inner validation subset of the official training portion was retained for model selection/early stopping; therefore, these results are not intended as a direct reproduction of published benchmark protocols that train on the complete official training partition.}

\section{Discussion}

\subsection{Classical Machine Learning vs. Deep Learning}
The results show that the relative performance of classical machine learning and deep learning depended on the dataset. On UCI-HAR, the deep learning models achieved slightly higher macro-F1 values than the classical machine learning models under the common ACC+GYRO setting. CNNLSTM obtained 0.936, followed by CNNBiLSTMAttn at 0.935, while the best classical machine learning model, LightGBM, achieved 0.929. However, the standard deviations of these configurations overlap considerably (0.031, 0.034, and 0.021, respectively), so the observed ordering on UCI-HAR should not be over-interpreted as a meaningful difference. In contrast, the pattern was different on KU-HAR. LightGBM achieved 0.857, while the best deep learning model, CNNLSTM, achieved 0.828. SVM and Random Forest also reached 0.847 and 0.836, respectively, exceeding several of the evaluated deep learning models. These results indicate that deep learning did not consistently outperform classical machine learning under the evaluated setting. The difference between the two approaches may be related to how well handcrafted statistical features represent the activity patterns in a particular dataset, while deep learning models learn representations directly from raw sensor sequences. The results therefore support using both model families as meaningful baselines when evaluating sensor-based HAR rather than assuming that increased model complexity necessarily leads to better performance. The results also highlight the importance of dataset characteristics. UCI-HAR produced relatively similar performance across the four deep architectures, whereas KU-HAR showed larger differences between models and greater variability across folds. Thus, model selection in sensor-based HAR should consider both predictive performance and the characteristics of the target dataset.

\subsection{Effect of Sensor Modality}
The results indicate that sensor modality has a substantial effect on activity recognition performance. Across the models evaluated with all three modalities, the combined ACC+GYRO modality generally produced higher macro-F1 than either ACC or GYRO alone on both datasets. On KU-HAR, combining ACC and GYRO increased macro-F1 for all five evaluated models. For example, LightGBM increased from 0.787 with ACC and 0.796 with GYRO to 0.857 with ACC+GYRO, while CNNBiLSTMAttn increased from 0.788 with ACC to 0.801 with the combined modality. A similar pattern was observed on UCI-HAR, where ACC+GYRO consistently produced higher macro-F1 than GYRO alone and generally exceeded ACC alone. The improvement was particularly pronounced for the classical feature-based models, whereas CNNBiLSTMAttn showed the smallest gain among the evaluated configurations, likely because its performance was already high with ACC alone. However, the contribution of each individual modality was dataset and model-dependent. On KU-HAR, GYRO slightly outperformed ACC for LightGBM and Random Forest, whereas CNNBiLSTMAttn performed better with ACC. On UCI-HAR, ACC was consistently more informative than GYRO for the models included in the modality analysis. Overall, the findings suggest that accelerometer and gyroscope signals provide complementary information for the evaluated HAR tasks. However, the usefulness of an individual sensor type depends on both the dataset and the model, so modality selection should be considered jointly with the characteristics of the recognition task.

\subsection{Effect of Evaluation Protocol}
The evaluation protocol had a substantial effect on reported performance. The primary evaluation used five-fold StratifiedGroupKFold, with subjects as groups for UCI-HAR and participants as groups for KU-HAR, preventing the same group from appearing in both training and test folds. The grouped and random split comparison was performed using a single representative configuration, CNNBiLSTMAttn with ACC+GYRO. On UCI-HAR, macro-F1 increased from 0.9346 under grouped evaluation to 0.9672 under random splitting, a difference of 0.0326. On KU-HAR, the increase was larger, from 0.8014 to 0.8944, corresponding to a difference of 0.0930. The random split ignores group identities and was included specifically as a leakage ablation, whereas the grouped protocol was used as the primary evaluation setting. The larger increase observed on KU-HAR may be related to differences in dataset structure and participant-level data distribution; however, the present ablation does not isolate the contribution of individual factors. The standard deviation also decreased on KU-HAR, from 0.107 under grouped evaluation to 0.031 under random splitting, indicating lower fold-to-fold variability under the random protocol. These results show that the choice of data-splitting strategy can influence measured performance by a margin comparable to, or larger than, the differences commonly observed between competing model architectures. Therefore, comparisons based only on random splits should be interpreted carefully, particularly when multiple samples from the same subject or participant are available, and the evaluation protocol should be reported explicitly when presenting and comparing HAR results.

\subsection{Dataset Characteristics and Error Patterns}
\label{sec:disc_error}
The two datasets exhibited different recognition characteristics. UCI-HAR contains six activity classes, whereas KU-HAR contains 18 classes with more diverse and fine-grained activity categories, including several with similar motion patterns. The error patterns differed accordingly. On KU-HAR, the lowest recalls were observed for Stair-up (0.398), Stair-down (0.407), and Run (0.555), indicating substantial confusion among these dynamic activities. On UCI-HAR, the weakest recalls were STANDING (0.822) and SITTING (0.881), while the three WALKING classes reached recalls of 0.965 or higher. The errors were therefore concentrated around similar static postures rather than across dynamic activities. The larger standard deviation on KU-HAR (0.107 for CNNBiLSTMAttn with ACC+GYRO, compared with 0.034 on UCI-HAR) is largely driven by a single fold. In this fold, all evaluated models showed a sharp drop in macro-F1 (approximately 0.53--0.64 versus 0.79--0.91 in the other folds). Inspection of the group distribution showed that one participant contributed approximately 1900 windows, nearly ten times the median of 180 windows. When this participant was placed in the held-out fold, the remaining training groups did not cover the corresponding motion patterns, producing a substantial drop in macro-F1. These results indicate that dataset characteristics, including the number and type of activity classes and the distribution of data across participants, are important when interpreting HAR performance. The lower mean and larger variance on KU-HAR are therefore not attributable to model architecture alone. They reflect the greater diversity of the recognition task, the class imbalance shown in Figure~\ref{fig:datasets}, and the uneven per-participant data volume that produced the fold-level heterogeneity discussed above.

\section{Limitations and Future Directions}

\subsection{Single Seed}
The experiments were conducted using a single random seed (42) with five grouped cross-validation folds. The reported fold-level variability therefore captures variation across data partitions but does not reflect variation across random initializations or training seeds. Consequently, the reported performance differences between model configurations should be interpreted as indicative rather than as evidence of robustness across independent training runs. Future work should evaluate additional seeds, such as 42, 43, and 44, to assess the robustness of the reported results across training runs.

\subsection{Fixed Hyperparameters and Model Scope}
All models were evaluated using hyperparameters fixed a priori and applied consistently across datasets and sensor modalities. No hyperparameter tuning was performed. This design provides a controlled comparison of model families, representations, and evaluation protocols, but it also limits conclusions about the maximum achievable performance of each model. In particular, the deep learning models were limited to four architectures with parameter counts between 37k and 316k (CNN1D, BiLSTM, CNNLSTM, and CNNBiLSTMAttn) and did not include more complex architectures such as Transformers. Therefore, the finding that deep learning did not consistently outperform classical machine learning should be interpreted within the evaluated model architectures and fixed hyperparameter settings, rather than as a general conclusion about all deep learning approaches.

\subsection{Partial Modality Grid}
The modality evaluation did not cover all model-modality combinations. The full model comparison (Table~\ref{tab:overall}) was conducted using the common ACC+GYRO setting, while the modality analysis included all three modalities for the classical machine learning models and CNNBiLSTMAttn among the deep learning models. CNN1D, BiLSTM, and CNNLSTM were evaluated only with ACC+GYRO. This design was chosen to keep the total number of trainings tractable on the available hardware while retaining a representative deep learning architecture for modality analysis. However, it prevents direct comparison of modality effects across all deep architectures. Therefore, the observed modality trends should not be generalized to CNN1D, BiLSTM, or CNNLSTM without additional experiments. Future work should evaluate the complete model-modality grid to provide a more comprehensive analysis of the interaction between architecture and sensor modality.

\subsection{Limited Leakage Ablation}
The grouped versus random split ablation was limited to CNNBiLSTMAttn with ACC+GYRO on both datasets. This configuration was selected for the protocol comparison, while the full experimental grid was evaluated only under the primary grouped setting. The ablation quantifies the effect of the evaluation protocol for one representative configuration, rather than providing a comprehensive leakage analysis across all architectures and modalities. The observed increases in macro-F1 (0.0326 on UCI-HAR and 0.0930 on KU-HAR) are specific to this configuration and should not be generalized to other models or sensor modalities. A more comprehensive leakage analysis across all model--modality combinations remains a direction for future work.

\subsection{Dataset-Specific Group and Distribution Characteristics}
Because complete participant groups were kept within individual folds, fold sizes varied substantially on KU-HAR. The folds contained between 2{,}812 and 5{,}413 test windows, with fold 2 containing the largest test set. This uneven distribution was reflected in the fold-level performance. The fold with the largest test set (fold 2, 5{,}413 windows) also showed the lowest macro-F1 (0.6154 for CNNBiLSTMAttn with ACC+GYRO), while the fold with the smallest test set (fold 1, 2{,}812 windows) showed one of the highest (0.8716). The greater variability observed on KU-HAR may partly reflect the uneven distribution of participant data across folds. The five-fold grouped protocol provides a group-independent evaluation, but its estimate remains influenced by how participant groups are distributed across folds. Future work could investigate alternative strategies, such as leave-one-group-out or a larger number of folds, to obtain a finer assessment of performance variability.

\subsection{Class Imbalance and Small-Sample Classes}
The datasets contain unequal numbers of samples across activity classes, particularly in KU-HAR, where class sizes range from a few hundred to nearly two thousand windows per class. The experiments addressed this imbalance through the use of macro-averaged precision, recall, and F1-score, which give equal weight to each class during evaluation. No resampling or class-weighting strategy was applied. This design ensures that the reported macro-F1 reflects performance across all activity classes rather than being dominated by frequent ones. However, classes with fewer samples are expected to show higher estimation variability. It should be noted that the classes with the lowest recall in Section~\ref{sec:disc_error} (Stair-up 0.398, Stair-down 0.407, Run 0.555) are not the smallest classes in KU-HAR, indicating that class size alone does not explain the observed error patterns. Other factors, such as similarity between motion patterns, likely contribute more to these errors. Future work could investigate class-aware training strategies, such as resampling or class-weighted objectives, and evaluate their effect on minority-class recognition.

\section{Conclusion}
This study presented a controlled comparative evaluation of classical machine learning and deep learning approaches for sensor-based human activity recognition on UCI-HAR and KU-HAR. The experiments considered different model families, sensor modalities, and evaluation protocols using grouped cross-validation to prevent subject- or participant-level overlap between training and test folds. The results show that deep learning did not consistently outperform classical machine learning. On UCI-HAR, the evaluated deep learning models achieved slightly higher macro-F1 values, while on KU-HAR, LightGBM achieved a higher macro-F1 than the evaluated deep learning models under the common ACC+GYRO setting. Sensor modality also affected performance. Combining accelerometer and gyroscope signals generally produced higher macro-F1 than using either modality alone, although the contribution of each sensor modality varied across datasets and models. The evaluation protocol had a substantial effect on the measured performance. Random splitting increased macro-F1 by 0.0326 on UCI-HAR and 0.0930 on KU-HAR compared with grouped evaluation for the analyzed configuration, highlighting the importance of group-aware evaluation in sensor-based HAR. Overall, the findings indicate that HAR performance depends not only on model architecture, but also on sensor modality, dataset characteristics, and evaluation protocol. These results support the use of controlled, group-aware comparisons when assessing HAR methods and provide a basis for future studies involving broader model--modality combinations, multiple random seeds, and more extensive model optimization.

\bibliographystyle{unsrt}
\bibliography{references}

\end{document}